\documentclass[a4paper, 10pt, conference]{ieeeconf}      % Use this line for a4
\usepackage{FG2020}
\usepackage{amssymb}
\FGfinalcopy % *** Uncomment this line for the final submission

\IEEEoverridecommandlockouts                              % This command is only
\usepackage{bm}
\usepackage{graphicx}
\usepackage{amsmath}
\usepackage{dblfloatfix}
\usepackage{float}
\usepackage{multirow}
\usepackage{subfig}
\usepackage{array}

\def\FGPaperID{7} % *** Enter the FG2020 Paper ID here

\title{\LARGE \bf
Head2Head: Video-based Neural Head Synthesis
}

\author{\parbox{16cm}{\centering
    {\large Mohammad Rami Koujan$^{*1,3}$, Michail Christos Doukas$^{*2,3}$, Anastasios Roussos$^{4,3}$, Stefanos Zafeiriou$^{2,3}$}\\
    {\normalsize
    $^{1}$College of Engineering, Mathematics and Physical Sciences, University of Exeter, UK\\  $^{2}$Department of Computing, Imperial College London, UK\\$^{4}$Institute of Computer Science (ICS), Foundation for Research and Technology - Hellas (FORTH), Greece\\ $^{3}$FaceSoft.io, London, UK\\}}
    \thanks{$^{*}$ Equal contribution.}% <-this % stops a space
}

\begin{document}

\ifFGfinal
\thispagestyle{empty}
\pagestyle{empty}
\else
\author{Anonymous FG2020 submission\\ Paper ID \FGPaperID \\}
\pagestyle{plain}
\fi
\maketitle

%%%%%%%%%%%%%%%%%%%%%%%%%%%%%%%%%%%%%%%%%%%%%%%%%%%%%%%%%%%%%%%%%%%%%%%%%%%%%%%%
\begin{abstract}
In this paper, we propose a novel machine learning architecture for facial reenactment. 
In particular, contrary to the model-based approaches or recent frame-based methods that use Deep Convolutional Neural Networks (DCNNs) to generate individual frames, we propose a novel method that (a) exploits the special structure of facial motion (paying particular attention to mouth motion) and (b) enforces temporal consistency. We demonstrate that 
the proposed method can transfer facial expressions, pose and gaze of a source actor to a target video in a photo-realistic fashion more accurately than state-of-the-art methods. 

\end{abstract}
%%%%%%%%%%%%%%%%%%%%%%%%%%%%%%%%%%%%%%%%%%%%%%%%%%%%%%%%%%%%%%%%%%%%%%%%%%%%%%%%
\section{INTRODUCTION}

%Facial reenactment aims at transferring the expression of a source actor to a target face. This is a challenging problem, in the cutting edge of today’s research and technology, with many applications that range from video editing, visual effects and movie dubbing to telepresence and virtual reality. Recent works have been able to produce impressive results that have attracted an increased attention by the research community, the industry, as well as the general public. 

Facial reenactment aims at transferring the expression of a source actor to a target face. It is a challenging problem, in the cutting edge of research and technology, with many applications in video editing, movie dubbing, telepresence and virtual reality. Recent works have produced impressive results and have attracted the attention of the research community, the industry, as well as the general public.

%The vast majority of facial reenactment methods transfer the expressions of the source actor by modifying the deformations within solely the inner facial region of the target actor, without altering the head motion or other elements of the target video \cite{Liu2001, Olszewski2017, Suwajanakorn2017, thies2015realtime, face2face, Vlasic2005}. Thus, even in cases where this expression transfer is performed in an especially realistic way, the overall reenactment result might seem uncanny and non-plausible, as the head and upper-body motion of the target do not really match with the transferred expressions. 

The majority of facial reenactment methods transfer the expressions of the source actor by modifying the deformations within solely the inner facial region of the target actor, without altering the head movements of the target video \cite{Liu2001, Olszewski2017, Suwajanakorn2017, thies2015realtime, face2face, Vlasic2005}. Thus, even in cases where this expression transfer is performed well, the overall reenactment result might seem uncanny and non-plausible, as the head motion of the target may not match with the transferred expressions. 
%To tackle these problems, a couple of very recent methods attempt to transfer not only the facial expressions, but also the full 3D head motion \cite{elor2017bringingPortraits, deepvideoportraits}. This is much more challenging, since it requires that occluded parts of the background and the face are synthesised realistically. However, despite the promising results, these methods have limitations in terms of the realism of the reenactment result. Averbuch-Elor et al.~\cite{elor2017bringingPortraits} use a target portrait photo instead of a video and cannot deal well with large variations in the head pose, the target gaze is not controlled and the mouth appearance is copied from the source actor. The approach of Kim et al. ~\cite{deepvideoportraits} performs complete motion transfer from a source to a target actor. Nonetheless, there are cases where the head movements and facial expressions in the generated video do not follow closely the ones in the source sequence. Additionally, the interior of the mouth occasionally appears unrealistic and temporally inconsistent between subsequent frames.
Various recent methods attempt to transfer the entire head motion \cite{X2Face, deepvideoportraits, elor2017bringingPortraits, fewshot}. Despite their promising results, these methods have limitations in terms of how realistic the reenactment result looks. \textit{X2face} \cite{X2Face} is a warping-based approach, which is sometimes unsuccessful on preserving the identity of target person and the results often look unrealistic. \textit{Deep Video Portraits} ~\cite{deepvideoportraits} performs image-based head reenactment, by conditioning an image-to-image translation neural network on 3D facial information. As frames are synthesised independently from each other, there are cases where videos exhibit temporal incoherence between the generated frames, especially in the mouth region.

%In addition, Kim et al. ~\cite{deepvideoportraits} do not adapt the eye geometry of the source actor to the target, which leads to unsuccessful synthesis in case of dissimilar facial structures. Furthermore, there are cases where the results of \cite{deepvideoportraits} exhibit temporal incoherence between the generated frames of the reenacted target video.

%Our proposed approach overcomes all the aforementioned limitations. We fully transfer the human head characteristics (pose, facial expression, eye gaze movement) from a source to target video, % This is tackled in our framework while while preserving the identity of the target and maintaining a consistent motion of the upper body part and the background, which are present in the original video. Considering that the affinity of humans for a synthetic video reduces significantly even for minor mistakes in the appearance and motion (uncanny valley effect), we give specific attention in the details of the mouth and teeth. Our system generates photo-realistic and temporally consistent videos of faces. 

Our proposed approach overcomes all the aforementioned limitations. We fully transfer the pose, facial expression and eye gaze movement from a source to target video, while preserving the identity of the target and maintaining a consistent motion of the upper body part. Given that people easily detect mistakes in the appearance of a human face (uncanny valley effect), we give special attention in the details of the mouth and teeth. Our system generates photo-realistic and temporally consistent videos of faces. 

\begin{figure*}[t!]
    \centering
    \includegraphics[scale=0.42]{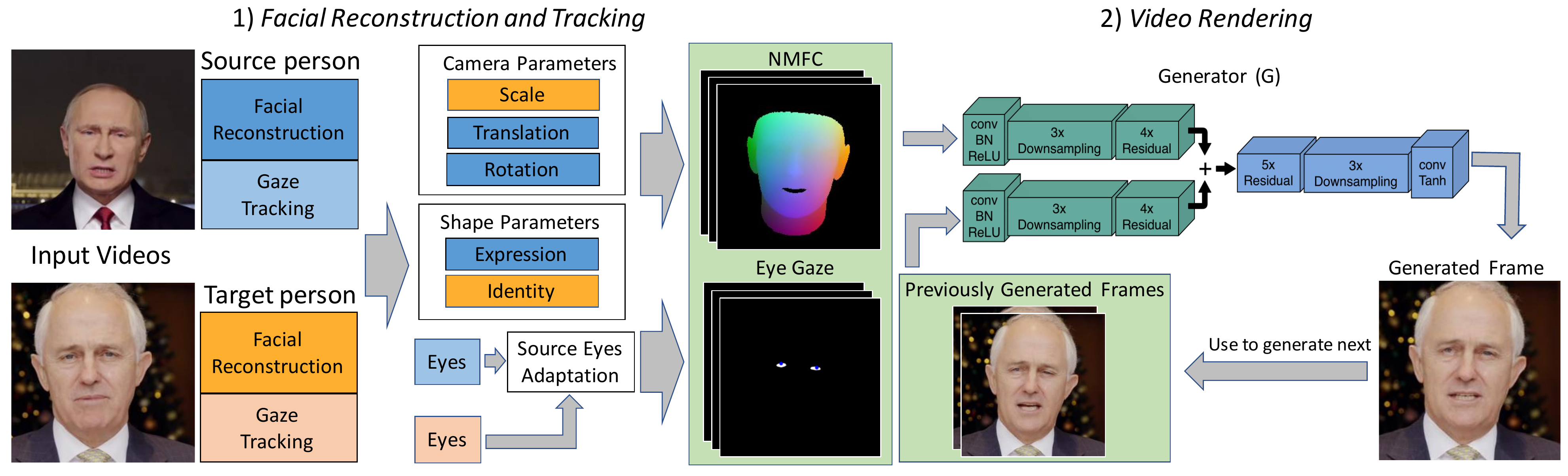}
    \caption{The pipeline of our Head2Head approach marked by two subsequent stages: 1) facial reconstruction and tracking, 2) video rendering. The $\mathbf{NMFC}$ and eye gaze sequences computed in the first stage are used to drive the video synthesis. During training, both the source and target frames come from the same (target) video, since we perform self-reenactment.}
    \label{fig:fig1}
\end{figure*}

\section{Related Work}

\subsection{3D Face Reconstruction}

Human faces have attracted substantial attention in the computer vision field due to their centrality in many applications. A growing line of research revolves around reconstructing the 3D geometry of the face and employing this information in further ad-hoc steps, as we do in this work. Some attempts on 3D face reconstruction \cite{barron2014shape, snape2015automatic, snape2014kernel}, known as Shape From Shading, work on approximating the image formation process and postulate simplified assumptions about the lighting and illumination models leading to the formation of the image, while others, known as Structure from Motion (SfM), capitalise on the geometric constraints in multiple images of the same object to approach the reconstruction task \cite{garg2013dense, grasshof2017projective, hernandez2017accurate}. 3D Morphable Models (3DMMs) have been researched and used substantially in the literature since the pioneering work of Blanz and Vetter \cite{blanz1999morphable}, with many extensions \cite{koujan2018combining, booth20183d, li2017learning}. 3DMMs are generative parametric models for the 3D representation of human faces. They are built from a set of 3D facial scans, coupled to each other with anatomical correspondences, and can represent any unseen faces as a linear combination of the training set.

\subsection{Facial Reenactment and Full Head Reenactment Methods}

A plethora of works is devoted to the problem of facial reenactment \cite{Olszewski2017, thies2015realtime, face2face, elor2017bringingPortraits}. Expressions are transferred either by 2D warping techniques, based on dense motion fields \cite{Liu2001, Garrido2014, elor2017bringingPortraits}, or by fitting and manipulating parametric 3D face models \cite{Vlasic2005, thies2015realtime, face2face}. A well-known facial reenactment system is \textit{Face2Face} \cite{face2face}, which relies on monocular 3D face reconstruction on both the source and target videos and operates in real time. \textit{Neural Textures} \cite{DeferredNeuralRendering} is another approach to facial reenactment, achieving high quality results. In this method, texture features are learned from the target video and translated to RGB images with a neural renderer. 

Although many of these facial reenactment systems produce highly realistic results, they do not provide complete control over the generated video. There is a limited number of works in the direction of full head motion transfer \cite{X2Face, deepvideoportraits, elor2017bringingPortraits, fewshot}. Averbuch-Elor et al.~\cite{elor2017bringingPortraits} use a target portrait photo instead of a video and cannot deal well with large variations in the head pose. The target gaze is not controlled and the mouth appearance is copied from the source actor. Zakharov \emph{et al.} \cite{fewshot} propose a network that synthesises frames conditioned on landmarks, using a few-shot adversarial learning approach. However, their method does not adapt the landmarks extracted from the source actor to the head geometry of the target, leading to identity mismatches of the generated video. 
%As far as we are aware, \textit{Deep Video Portraits} \cite{deepvideoportraits} is the only approach that performs full motion transfer from the source to the target. However, there are cases where the mouth interior of the target actor exhibits temporal incoherence along consecutive frames, as well as instances of mismatching between the source and target actor's expressions and poses.

\subsection{Image and Video Synthesis}

%Generative Adversarial Networks (GANs) \cite{ganGoodfellow} are widely used for the task of photo-realistic image and video synthesis. Instead of conditioning synthesis on noisy distributions \cite{ganGoodfellow, PGAN}, other data types, such as class labels \cite{cgan} or images \cite{pix2pix, CycleGAN2017, pix2pixHD}, are frequently used as input to the generator. There is a plethora of works on image-to-image translation \cite{UNIT, TaigmanPW16}, with impressive results in domain and style transfer. Various works have extended the GAN framework on video synthesis applications, such as unconditional video generation \cite{Saito2017TemporalGA, VondrickNIPS2016, tulyakov2017mocogan}, future prediction \cite{Denton2017, Finn2016} and style transfer \cite{Chen_2017_ICCV}. In this work, we exploit a GAN-based approach for rendering photo-realistic video frames, which provides temporal stability and paramount image quality.

Generative Adversarial Networks (GANs) \cite{ganGoodfellow} have been widely used for photo-realistic image and video synthesis. Instead of conditioning synthesis on noisy distributions \cite{ganGoodfellow, PGAN}, other data types are frequently used as input to the generator, such as class labels \cite{cgan} or images \cite{pix2pix, CycleGAN2017, pix2pixHD}. There is extended research on image-to-image translation \cite{UNIT, TaigmanPW16, pix2pix, pix2pixHD}. Other methods \cite{Saito2017TemporalGA, tulyakov2017mocogan, Denton2017, Chen_2017_ICCV} utilise the GAN framework on video synthesis tasks, such as \textit{vid2vid} \cite{vid2vid}. In this work, we exploit a GAN-based approach for rendering photo-realistic video frames, providing temporal stability and paramount image quality.
%The neural network proposed in Deep Video Portraits \cite{deepvideoportraits} follows an image translation approach. On the contrary, our method is more closely related to video-to-video translation.  Our system generates photo-realistic and temporally consistent videos of faces, following ideas on video translation from vid2vid \cite{vid2vid}.

\section{Methodology}
%Our proposed approach addresses the highly-challenging problem of head to head video translation. More specifically, we fully transfer the human head characteristics (pose, facial expression, eye gaze movement) from a source to target actor's video. This is tackled in our framework while preserving the identity of the target and maintaining a consistent motion with the upper body part and the background present in the original video. Our \textbf{head2head} pipeline is shown in Fig. ??. Two main consecutive stages are adopted: 1) facial reconstruction and tracking, 2) learning-based video rendering network. 

Our proposed approach addresses the highly-challenging problem of full-head reenactment. More specifically, we fully transfer head pose, facial expression and eye gaze movement from a source to target actor's video. Our \textbf{head2head} pipeline, illustrated in Fig. \ref{fig:fig1}, consists of two main and consecutive stages: 1) 3D facial reconstruction and tracking, and 2) learning-based video rendering, using a neural network.

\subsection{Facial Reconstruction and Tracking}
To effectively disentangle the human head characteristics in a transferable and photo-realistic way between different facial videos, we benefit from the prior knowledge about the targeted object. This knowledge has been effectively addressed and captured with 3DMMs \cite{blanz1999morphable}. Thus, we harness the power of 3DMMs for 3D reconstructing and tracking the faces appearing in the input sequences. Given a sequence of $T$ frames $\mathcal{F}_{1:T}= \{f_t \mid {t=1,\dots,T}\}$, the 3D reconstruction and tracking stage produces two sets of parameters: 1) shape parameters $\mathcal{S}=\{\bm{s}_t \mid \bm{s}_t \in \mathbb{R}^{n_i+n_e}, t=1,...,T\}$ , and 2) camera parameters $\mathcal{P}=\{\bm{p}_t \mid \bm{p}_t \in \mathbb{R}^{7}, t=1,...,T\}$, embodying rotation, translation and orthographic scale.

\textbf{Shape representation.} With 3D Morphable Models, a 3D facial shape $\mathbf{x}_t=[x_1, y_1, z_1,..., x_N, y_N, z_N]^T \in \mathbb{R}^{3N}$ can be represented mathematically as:
\begin{equation}
    \label{eq:3DMM}
\mathbf{x}(\bm{s}_t^i,\bm{s}_t^e)=\bar{\mathbf{x}}+\mathbf{U}_{id} \bm{s}_t^i+ \mathbf{U}_{exp} \bm{s}_t^e
\end{equation}
where $\mathbf{\bar{x}}\in \mathbb{R}^{3N}$ is the overall mean shape vector of the morphable model, given by $\mathbf{\bar{x}}=\mathbf{\bar{x}}_{id}+\mathbf{\bar{x}}_{exp}$,  with $\mathbf{\bar{x}}_{id}$ and $\mathbf{\bar{x}}_{exp}$ depicting the mean identity and expression of the model, respectively. $\mathbf{U}_{id} \in \mathbb{R}^{3N\times n_i}$ is the identity orthonormal basis with $n_i$ principal components ($n_i\ll3N$), $\mathbf{U}_{exp} \in \mathbb{R}^{3N\times n_e}$ is the expression orthonormal basis with the $n_e$ principal components ($n_e\ll3N$) and $\bm{s}_t^i \in \mathbb{R}^{n_i}$, $\bm{s}_t^e \in \mathbb{R}^{n_e}$ are the identity and expression parameters of the morphable model. We denote the joint identity and expression parameters of a frame $\bm{t}$ by $\bm{s}_t=[{\bm{s}_t^i}^T, {\bm{s}_t^e}^T]^T$.  In the adopted model \eqref{eq:3DMM}, the 3D facial shape $\mathbf{x}$ is a function of both identity and expression coefficients ($\mathbf{x}(\bm{s}_t^i, \bm{s}_t^e)$), with expression variations being effectively represented as offsets from a given identity shape.

\textbf{Video Fitting.} To estimate the shape and pose parameters of the source and target sequences in a quick and robust way, we follow a novel batch landmark-based approach that takes into account the information from all video frames simultaneously and exploits the rich dynamic information usually contained in facial videos. Similar to \cite{Deng2018}, we exploit the fact that the current state-of-the-art in facial landmarking can achieve highly-reliable landmark localisation and therefore fuse the landmarks information  with high-quality 3D face models, as the one described in \eqref{eq:3DMM},
to achieve robust and accurate 3D face reconstruction results. We assume that the camera performs scaled orthographic projection (SOP) and that the identity parameters $\bm{s}_t^i$ are fixed (but unknown) over all the frames, letting however the expression parameters $\bm{s}_t^e$ as well as the camera parameters (scale and 3D pose) to vary from one frame to another. 
In brief, for a given sequence of frames, we minimise a cost function that consists of three terms, see \eqref{eq: energy_eq} : \textbf{a)} a sum of squared 2D landmark reprojection errors over all frames ($E_{l}$),  \textbf{b)} a shape priors term ($E_{pr}$) that imposes a quadratic prior over the identity and per-frame expression parameters and \textbf{c)} a temporal smoothness term ($E_{sm}$) that enforces smoothness of the expression parameters in time, by using a quadratic penalty of the second temporal derivatives of the expression vector. 
\begin{equation}
    \label{eq: energy_eq}
    E(\mathcal{S}, \mathcal{P})= w_{l} E_{l}(\mathcal{S}, \mathcal{P}) + w_{pr} E_{pr}(\mathcal{S}) + w_{sm} E_{sm}(\mathcal{S}^e)
\end{equation}
In addition, to deal with outliers (e.g. frames with strong occlusions that cause gross errors in the landmarks), we also impose box constraints on the identity and per-frame expression parameters.
Assuming that the camera parameters ($\mathcal{P}$) in \eqref{eq: energy_eq} have been estimated in an initialisation stage, the minimisation of the cost function results in a large-scale least squares problem with box constraints, which we solve efficiently by using the reflective Newton method of \cite{coleman1996reflective}. More details about the initialisation stage of our video fitting step are available in the supplementary material.

\textbf{3DMM details.} For our set of experiments, the identity part of the 3DMM, $\{\mathbf{\bar{s}}_{id}, \mathbf{U_{id}}\}$, originates from the Large Scale Morphable Model (LSFM) \cite{booth20163d, booth2018large} built from approximately 10,000 scans of different people, the largest 3DMM ever constructed, with varied demographic information. In addition, the expression part of the model, $\{\mathbf{\bar{x}}_{exp}, \mathbf{U}_{exp}\}$ originates from the work of Zafeiriou et al.~\cite{Zafeiriou2017}, who built it using the blendshapes model of Facewarehouse \cite{cao2014facewarehouse} and adopting Nonrigid ICP \cite{cheng2017statistical} to register the blendshapes model with the LSFM model.

\textbf{Gaze tracking.} Since the human gaze direction is not captured generally by 3D face scanners, 3DMMs of facial shapes do not represent this characteristic. We, therefore, employ a state-of-the-art gaze tracking technique \cite{park2018deep} for tracking the eyes in the source and target sequences.

Many state-of-the-art approaches \cite{deepvideoportraits, face2face}, etc. rely on the analysis-by-synthesis framework for fitting 3DMMs to images, which requires estimating more parameters (e.g. illumination and reflectance) and solving a highly ill-posed problem. On the other hand, our facial reconstruction and tracking stage is a sparse-landmarks-based fast approach, which requires only 68 facial landmarks extracted by \cite{guo2018stacked}, as well as the frame sequence. Thanks to our novel video rendering framework, the facial representation extracted by our face tracker encapsulates adequate information for synthesising photo-realistic and temporally smooth videos, removing the need for more elaborate and slower 3D facial reconstruction and tracking techniques. 

\subsection{Conditioning Images Generation}
%Training our video renderer requires:
%Our carefully designed video renderer expects as input: 
%1) a sequence composed of RGB images of the target actor and 2) the per-frame face parameterisation, extracted from that sequence, during the face reconstruction and tracking step.
%Such a parameterisation conditions the video renderer's results and plays a vital role during training and test time to learn and transfer the different head characteristics between dissimilar sequences smoothly.
%This parameterisation, which disentagles identity from expression, allows us to train our video rendering network on a specific target actor, and transfer the expression and pose of another source actor with different head characteristics, during test.

Given the estimated shape and camera parameters from both the source and target frames at time $t$, we replace the identity coefficients and the scale parameter of the source actor with the ones from the target, creating the "hybrid" shape and camera parameters $\bm{s}_t$, $\bm{p}_t$, as shown in Fig. \ref{fig:fig1}. Then, instead of feeding these per-frame face parameters directly to the video rendering network, we create a more meaningful representation in the image space. %Given the estimated head parameters, namely, shape ($\bm{s}_t$) and camera ($\bm{p}_t$), we rasterize the 3D reconstruction of this frame, producing a visibility mask ($\mathbf{M} \in \mathbb{R}^{W\times H}$) in the image space.
We rasterize the 3D facial shape $\mathbf{x}(\bm{s}_t)$, producing a visibility mask ($\mathbf{M} \in \mathbb{R}^{W\times H}$) in the image space. Each pixel of $\mathbf{M}$ stores the ID of the corresponding visible triangle on the 3D face from this pixel. Then, we encode the normalised x-y-z coordinates of the centre of this triangle in another image, termed as Normalised Mean Face Coordinates ($\mathbf{NMFC} \in \mathbb{R}^{W\times H \times3}$) image, and utilise it as conditional input of the video rendering stage, see equation (\ref{eq: NMFC}) below.
\begin{equation}
    \label{eq: NMFC}
    \mathbf{NMFC}_t= \mathcal{E}(\mathcal{R}(\mathbf{x}(\bm{s}_t), \bm{p}_t), \bar{\bm{x}}),
\end{equation}
where $\mathcal{R}$ is the rasterizer, $\mathcal{E}$ is the encoding function and $\bar{\bm{x}}$ is the normalised version of the utilised 3DMM mean face (see (\ref{eq:3DMM})), so that the x-y-z coordinates of this face $\in [0, 1]$. This representation is very convenient, as the rendering neural network learns to associate it with the corresponding RGB values, pixel by pixel, and, therefore, results in a realistic and novel video synthesis. 

In addition to the $\mathbf{NMFC}$ images, we condition the neural video renderer on the gaze images ($\mathbf{G}_t$), as can be seen in Fig. \ref{fig:fig1}. Gaze images are generated by connecting the extracted eye landmarks at the gaze tracking stage with edges, and filling the interior with color. The produced eye gaze frames are in exact correspondence with the NMFC frames, meaning that in both representations eyes should appear at the same pixel locations.

%This representation is more convenient for the rendering neural network to associate with the corresponding RGB frame, pixel by pixel, and, therefore, results in a realistic and novel video synthesis. 
%During the test time, the $\mathbf{NMFC}$ images can be created so that the parameters $\bm{S}_t^e$ and $\bm{p}_t$ can represent the facial expression and pose of a frame $t$ coming from a different sequence (source actor).

\subsection{Video Rendering Neural Network}

%We use a deep neural network for the task of video synthesis. Given the video of a source actor $\bm{y}_{1:T} = \{y_t \}_{t=1,\dots,T}$, first we compute the sequence of $\mathbf{NMFC}$ frames $\mathbf{NMFC}_{1:T}$ and the corresponding sequence of eye gaze frames $\mathbf{G}_{1:T}$. Then, the conditional input to our neural network $\bm{x_{1:T}}$ is computed, by concatenating these two frame sequences at the channel dimension, such that $\bm{x}_{1:T} \equiv \{x_t = (\mathbf{NMFC}_t, \mathbf{G}_t) \}_{t=1,\dots,T}$ and $x_t \in \R^{H \times W \times 6}$ for each time step $t$. 
Given a sequence of $\mathbf{NMFC}$ frames $\mathbf{NMFC}_{1:T}$ and the corresponding sequence of eye gaze frames $\mathbf{G}_{1:T}$,
the neural network learns to translate the conditional input video $\bm{x}_{1:T} \equiv \{x_t = (\mathbf{NMFC}_t, \mathbf{G}_t) | x_t \in \mathcal{R}^{H \times W \times 6} \}_{t=1,\dots,T}$  to a highly realistic and temporally coherent output video $\bm{\tilde{y}_{1:T}}$, which shows the target actor performing exactly the same head motions and eye blinks as the actor in the source video. We train this network in a self-reenactment setting, where the source actor coincides with the target actor. Therefore, the generated video $\bm{\tilde{y}}_{1:T}^i$ should be a reconstruction of the ground truth $\bm{y}_{1:T}^i$, which is considered both as the source and target video. 
We adopt a GAN framework for video translation, where the generator $G$ is trained in an adversarial manner, alongside an image discriminator $D_I$ and a multi-scale dynamics discriminator $D_D$, which ensures that the generated video is realistic, temporally coherent and conveys the same dynamics of the target video. We further improve the visual quality of the mouth area, by designing a dedicated mouth discriminator $D_M$.

\textbf{Generator $G$.} In order to model the dependence of the produced frames on previous video time steps, we condition synthesis of the $t$-th frame $\tilde{y}_t$ not only on the conditional input $x_t$, but also on the previous inputs $x_{t-1}$ and $x_{t-2}$, as well as the previously generated frames $\tilde{y}_{t-1}$ and $\tilde{y}_{t-2}$, thus:
\begin{equation}
    \tilde{y}_{t} = G(\bm{x}_{t-2:t}, \bm{\tilde{y}}_{t-2:t-1} ; \theta_G).
\end{equation}

The generator is applied sequentially and the frames are produced one after the other, until the entire output sequence has been produced. 
The architecture of this network is inspired by \textit{vid2vid} \cite{vid2vid}. 
%Nonetheless, we found that learning the optical flow in the generator would not increase the quality of our results, therefore $G$ creates a hallucinated frame without predicting the optical flow. This makes the network faster. %Additionally, given the resolutions of the videos we manipulate, a coarse-to-fine generator design \cite{Wang2018} is not necessary. 
It consists of two identical encoding pipelines (see Fig. \ref{fig:fig1}), where the first one receives the concatenated $\mathbf{NMFC}$ and eye gaze images $\bm{x}_{t-2:t}$, while the second one takes in the two previously generated images $\bm{\tilde{y}}_{t-2:t-1}$. Their resulting features are first added and then passed through the decoding pipeline, which brings the output $\tilde{y}_{t}$ in a normalised [-1,+1] range, using a $\tanh$ activation.

\textbf{Image discriminator $D_I$ and mouth discriminator $D_M$.} Both of these networks learn to distinguish real frames from synthesised ones. During training, a random time step $t'$ in the range $[1,T]$ is uniformly sampled. Then, the real pair $(x_{t'}, y_{t'})$ and the fake one $(x_{t'}, \tilde{y}_{t'})$ are fed in $D_I$. Moreover, the corresponding mouth regions $(x_{t'}^m, y_{t'}^m)$ and $(x_{t'}^m, \tilde{y}_{t'}^m)$ are cropped and passed to $D_M$. In order to force these generators to create high-frequency details in local patches of the frames, we use a Markovian discriminator architecture (PatchGAN), as in \cite{pix2pix} and \cite{vid2vid}.

\textbf{Dynamics discriminator $D_D$.} The dynamics discriminator is trained to detect videos with unrealistic temporal dynamics. This network receives a set of $K=3$ consecutive real frames $\bm{y}_{t':t'+K-1}$ or fake  frames $\bm{\tilde{y}}_{t':t'+K-1}$ in its input, which were randomly drawn from the video. To be more precise, $D_D$ is not conditioned only on these short video clips of length $K$. Given the optical flow $\bm{w}_{1:T-1}$ of the ground truth video $\bm{y}_{1:T}$, the purpose of $D_D$ is to ensure that the flow $\bm{w}_{t':t'+K-2}$ corresponds to the given video clip. Therefore, the dynamics discriminator should learn to identify the pair $(\bm{w}_{t':t'+K-2}, \bm{y}_{t':t'+K-1})$ as real and the pair $(\bm{w}_{t':t'+K-2}, \bm{\tilde{y}}_{t':t'+K-1})$ as fake. In practice, we employ a multiple scale dynamics discriminator, which performs the task described above in three different temporal scales. The first scale receives sequences in the original frame rate. Then, the two extra scales are formed by choosing not subsequent frames in the sequence, but sub-sampling the frames by a factor of $K$ for each scale.

\textbf{Objective function:} The objective of our GAN-based framework can be expressed as an adversarial loss. We use the LSGAN \cite{lsgan} loss, thus the adversarial objective of the generator is:
\begin{equation}
\begin{split}
& \mathcal{L}_{adv}^G = \dfrac{1}{2} \mathbb{E}_{t' \sim \mathcal{U}\{1,T-2\}}[(D_D(\bm{w}_{t':t'+1}, \bm{\tilde{y}}_{t':t'+2}) - 1)^2] \\
& + \dfrac{1}{2} \mathbb{E}_{t' \sim \mathcal{U}\{1,T\}}[(D_I(x_{t'}, \tilde{y}_{t'}) - 1)^2 + D_M(x_{t'}^{m}, \tilde{y}_{t'}^{m}) - 1)^2].
\end{split}
\label{eq:1}
\end{equation}
We add two more losses in the learning objective function of the generator: 1) a VGG loss $\mathcal{L}_{vgg}^G$ and 2) a feature matching loss $\mathcal{L}_{feat}^G$ \cite{pix2pixHD, vid2vid}, which is based on the discriminators. Given a ground truth frame $y_t$ and the synthesised frame $\tilde{y}_t$, we use the VGG network \cite{vgg} to extract visual features in different layers for both frames and compute the VGG loss as in \cite{pix2pixHD} and \cite{vid2vid}.
%Then, we compute the loss:
%\begin{equation}
%\mathcal{L}_{vgg}^G = \sum_i \dfrac{1}{H_i W_i C_i} || L^{(i)}(y_t) - L^{(i)}(\tilde{y}_t) ||_1,
%\end{equation}
%with $L^{(i)}$ being the $i$-th layer of VGG, with a feature map of size $H_i \times W_i \times C_i$. 
In a similar way, we compute the discriminator feature matching loss, by extracting features with the two discriminators $D_I$ and $D_D$ and computing the $\ell_1$ distance of these features for a fake frame $\tilde{y}_t$ and the corresponding ground truth $y_t$. The total objective for $G$ is given by:
\begin{equation}
\mathcal{L}^G = \mathcal{L}_{adv}^G + \lambda_{vgg} \mathcal{L}_{vgg}^G + \lambda_{feat}\mathcal{L}_{feat}^G  
\end{equation}
In order to balance out the loss terms above, we set $\lambda_{vgg} = \lambda_{feat} = 10$. The image and mouth discriminators as well as the dynamics discriminator are optimised under their corresponding adversarial objective functions. All discriminators share the same architecture, which is adopted from pip2pixHD \cite{pix2pixHD}. More details on our video rendering stage are in the supplementary material.

\begin{figure*}[h!]
    \centering
    \includegraphics[scale=0.53]{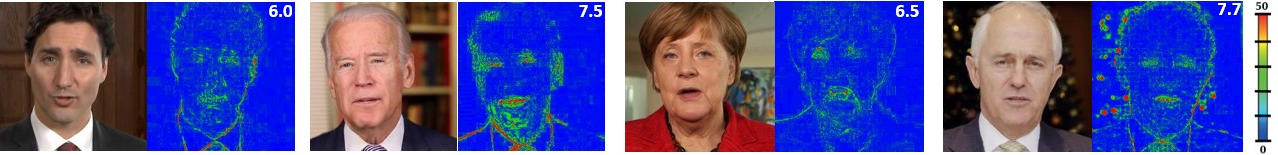}
    \caption{Quantitative assessment of the self-reenactment experiment on four different videos. In each pair, (left) synthesised frame, (right) corresponding error (heat) map. White numbers in the corner of the heat maps represent the average error for the entire corresponding test sequence. All images are in the range [0, 255].}
    \label{fig:quantitative_1}
\end{figure*}

\textbf{Optical facial flow.} %To generate as realistic dynamics as possible by our \textbf{head2head} network, it is very essential to estimate an accurate facial flow of the target video. 
To generate as realistic dynamics as possible by our \textbf{head2head} video rendering network, it is essential to condition the dynamics discriminator $D_D$ on a very accurate facial flow of the target video. 
Most state-of-the-art methods for optical flow estimation solve this problem without any prior assumptions about the objects appearing in consecutive images. Since human facial performances exhibit non-rigid and composite deformations as a result of very complex facial muscular interactions, capturing their flow using off-the-shelf state-of-the-art optical flow methods might not always produce visually convincing synthesis. To overcome this issue, we capitalise on the prior knowledge, as we target facial videos, and train a specific network for this task. To start with, we utilise a state-of-the-art network, called FlowNet2, for the optical flow estimation \cite{flownet2}.  This network was trained on publicly available images after rendering them with synthesised chairs modified by various affine transformations. We use the pretrained models of \cite{flownet2} and fine-tune their network on the 4DFAB dataset \cite{cheng20184dfab}, which comes with dynamic high-resolution 4D videos of subjects eliciting spontaneous and posed facial behaviours. To create the ground truth 2D flow, we use the provided camera parameters of the acquisition device and rasterize the 3D scans of around 750K frames so that the difference between each pair of consecutive frames represents the 2D flow. For the background, we generate the 2D flow estimates of the same 750K frames using the original FlowNet2 and use a masked End-Point-Error (EPE) loss so that the background flow stays the same and the foreground follows the ground truth 2D flow coming from the 4DFAB dataset.\\

\section{Experiments}

In this section, we demonstrate the capability of our \textbf{head2head} framework in transferring the full head pose, gaze, eye blinking and expression from a source to a target video.
%, as well as various combinations of these characteristics. 
Our approach was compared with state-of-the-art methods and achieved very competitive, visually appealing and realistic results. We conduct comprehensive experiments and ablation studies, probing the performance of our proposed approach both quantitatively and qualitatively. We collected a database from publicly-available videos, all having a spatial resolution of $256 \times 256$ pixels. This database consists of multiple subjects, mainly well-known politicians (see supplementary material for more results and visualisations). The \textbf{head2head} pipeline requires only a footage of a few minutes for training a model on the given target subject and takes only around 5 hours on an NVIDIA Titan V GPU to finish the training task. 
%The facial reconstruction and tracking step is highly parallelisable. For our set of experiments, we ran this step on four cores in parallel, achieving a rate of 16 fps for estimating the identity, expression and camera parameters. The eye gaze tracking step produces very quick estimations, as fast as 22 fps on an Intel Core i7 CPU. Once trained on a specific target video, our entire pipeline takes only 85ms to synthesise a frame with a fully reenacted head of the target according to the input source frame.  

\begin{figure}[h!]
    \centering
    \includegraphics[scale=0.2]{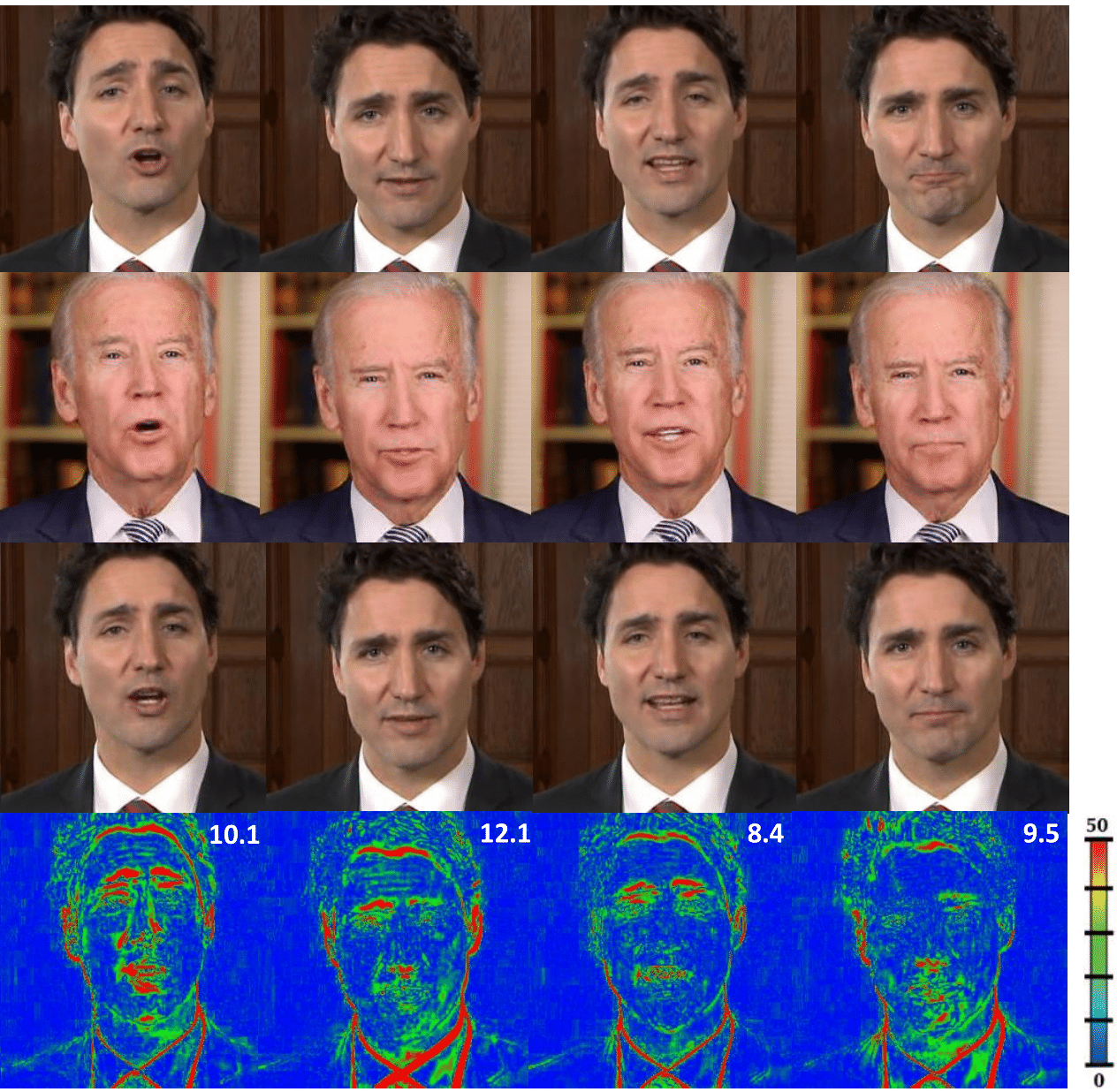}
    \caption{Cycle-head2head-reenactment results. Top to bottom: driving (source) sequence, manipulated target sequence, manipulated source sequence driven by the manipulated target sequence in the row before, per-pixel $\textit{l}_2$ distance between first and third row. Numbers in the top-right corners of error maps represent the average per-pixel error for each image.}
    \label{fig:quantitative_2}
\end{figure}

\subsection{Quantitative Results}
We quantitatively evaluate the performance of our method by conducting two sets of experiments: 1) self-reenactment, and 2) cycle-reenactment. For self-reenactment, each target video was split into a train (first two thirds) and a test (last third) set, and the average per-pixel error was computed over the test split. Note that the per-pixel error is defined as the $\ell_2$ distance between RGB images, assuming that the range of values in each color channel is from 0 to 255. Fig. \ref{fig:quantitative_1} demonstrates the obtained quantitative results on four %dissimilar videos 
different subjects (Justin, Joe, Merkel and Turnbull), with an average of $4K$ training frames per video. 

\begin{figure}[b!]
    \centering
    \includegraphics[width=1.0\linewidth]{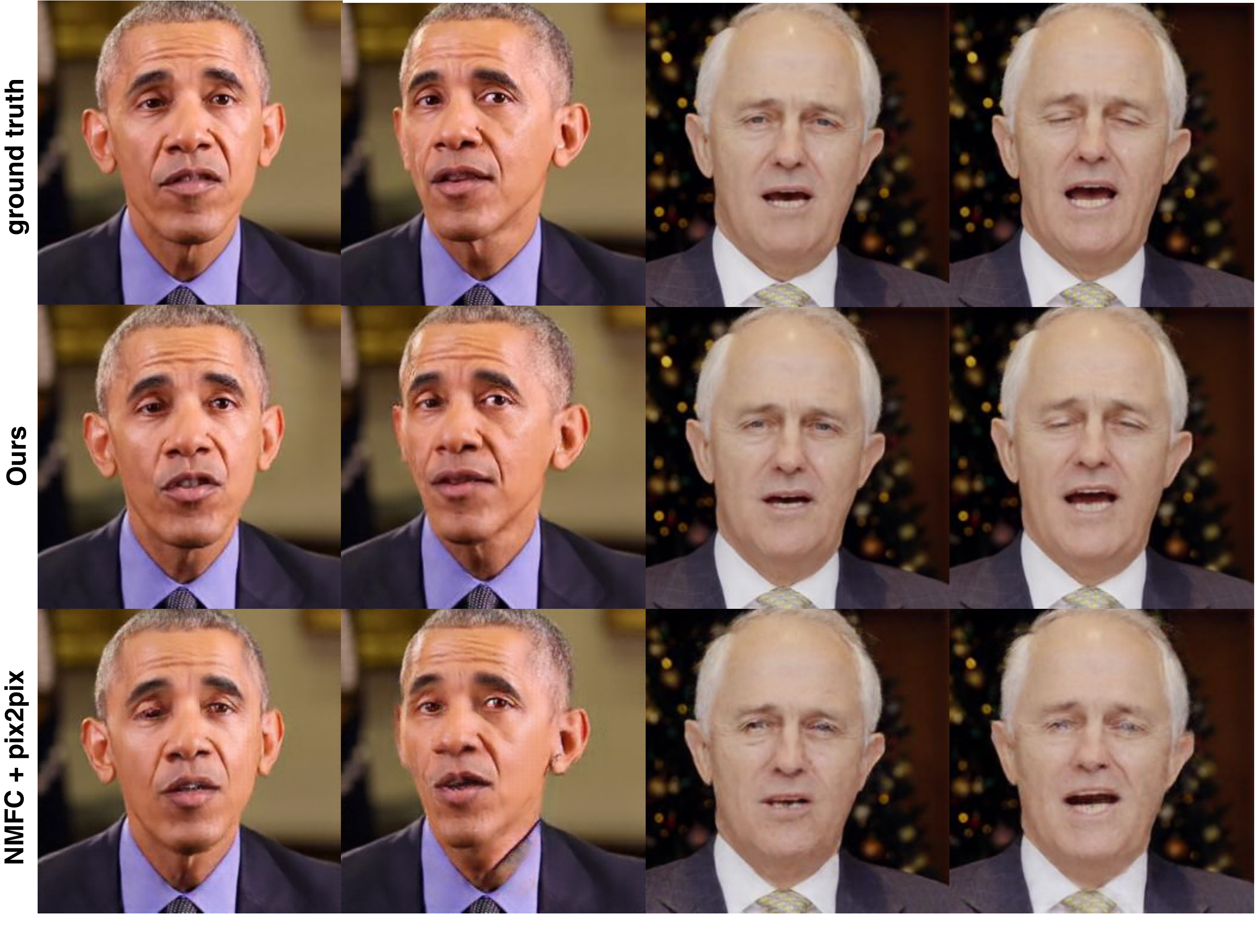}
    \caption{Comparison of our method with the \textit{pix2pix} \cite{pix2pix} baseline conditioned on $\mathbf{NMFC}$ images, on the self-reenactment task. We achieve better image quality and ground truth reconstruction. Please zoom in for details.}
    \label{fig:qualitative_2}
\end{figure}

%As a second experiment, we conducted a cycle-reenactment test to assess the performance of our approach in transparently transferring the human head attributes from a source to target and then back to the same source. More specifically, given a source sequence $\mathcal{X}$ and a target $\mathcal{Y}$, we first train a network $\mathcal{N}_{\mathcal{Y}}$ on the target $\mathcal{Y}$. Then, during test time, we transfer expression, pose and gaze direction from the source to the target, generating a video $\mathcal{N}_{\mathcal{Y}}(\mathcal{X})=\tilde{\mathcal{Y}}$. To complete the cycle, 1) another network $\mathcal{N}_{\mathcal{X}^{\prime}}$ was trained on the same source subject, but different frames so that $\mathcal{X} \cap \mathcal{X^{\prime} = \phi}$, 2) a fake video was generated  as $\mathcal{N}_{\mathcal{X}^{\prime}}(\tilde{\mathcal{Y}})=\tilde{\mathcal{X}}$, 3) the average per-pixel distance between $\mathcal{X}$ and $\mathcal{\tilde{X}}$ was computed. Fig. \ref{fig:quantitative_2} shows some randomly selected source-target-source ($\mathcal{X}\rightarrow \tilde{\mathcal{Y}} \rightarrow \tilde{\mathcal{X}}$) frames and the heat-maps (per-pixel $\textit{l}_2$ distance) between the frames in the first ($\mathcal{X}$) and third row ($\tilde{\mathcal{X}}$). The average per-pixel error over the entire test sequence, composed of 1K frames, is 9.3, given that generated images are in the range [0, 255]. 

As a second experiment, we conducted a cycle-reenactment test to assess the performance of our approach in transparently transferring the human head attributes from a source to a target and then back to the same source. More specifically, given a source sequence $\bm{X}$ and a target actor $\bm{Y}$, we first train a network $\mathcal{N}_{\bm{Y}}$ on the target $\bm{Y}$. Then, during test time, we transfer expression, pose and gaze direction from the source to the target, generating a video $\tilde{\bm{Y}} = \mathcal{N}_{\bm{Y}}(\bm{X})$. To complete the cycle, another network $\mathcal{N}_{\bm{X^{\prime}}}$ was trained on the same source subject of $\bm{X}$, using the training video $\bm{X'}$ with different frames from $\bm{X}$, such that $\bm{X} \cap \bm{X^{\prime}} = \emptyset$. Then, another fake video was generated, as $\tilde{\bm{X}} = \mathcal{N}_{\bm{X^{\prime}}}(\tilde{\bm{Y}})$ and the average per-pixel distance between $\tilde{\bm{X}}$ and $\bm{X}$ was computed. Fig. \ref{fig:quantitative_2} shows some randomly selected source-target-source ($\bm{X}\rightarrow \tilde{\bm{Y}} \rightarrow \tilde{\bm{X}}$) frames and the heat-maps (per-pixel $\ell_2$ distance) between the frames in the first ($\bm{X}$) and third row ($\tilde{\bm{X}}$). The average per-pixel error over the entire test sequence, composed of 1K frames, is 9.3.

\setcounter{figure}{5}

\begin{figure*}[b]
    \centering
    \includegraphics[scale=0.245]{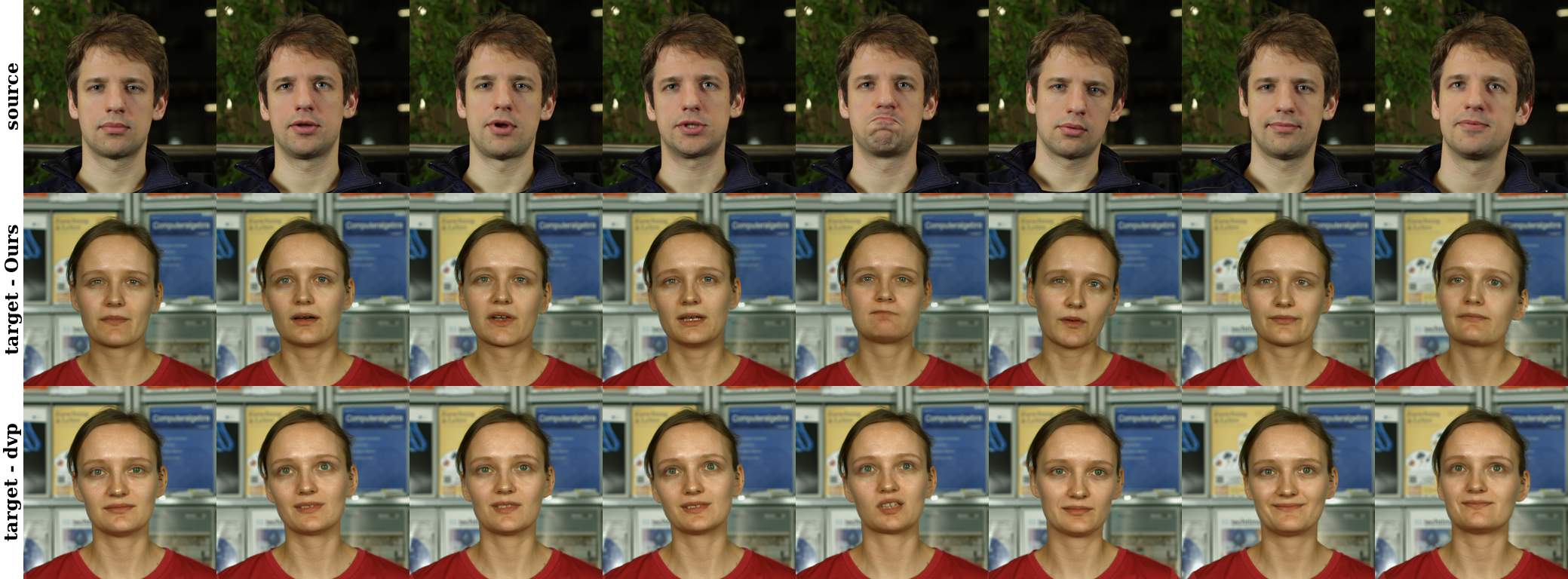}
    \caption{Comparison with \textit{Deep video Portraits} (dvp) \cite{deepvideoportraits}. Top to bottom: the driving (source) sequence, the target sequence generated with our \textbf{head2head} method, the corresponding frames with \textit{Deep video Portraits} method. Our poses and expressions match better to the source. Please zoom in for details.}
    \label{fig:qualitative_1}
\end{figure*}

\setcounter{figure}{4}

\begin{figure}[h!]
    \centering
    \includegraphics[width=0.86\linewidth]{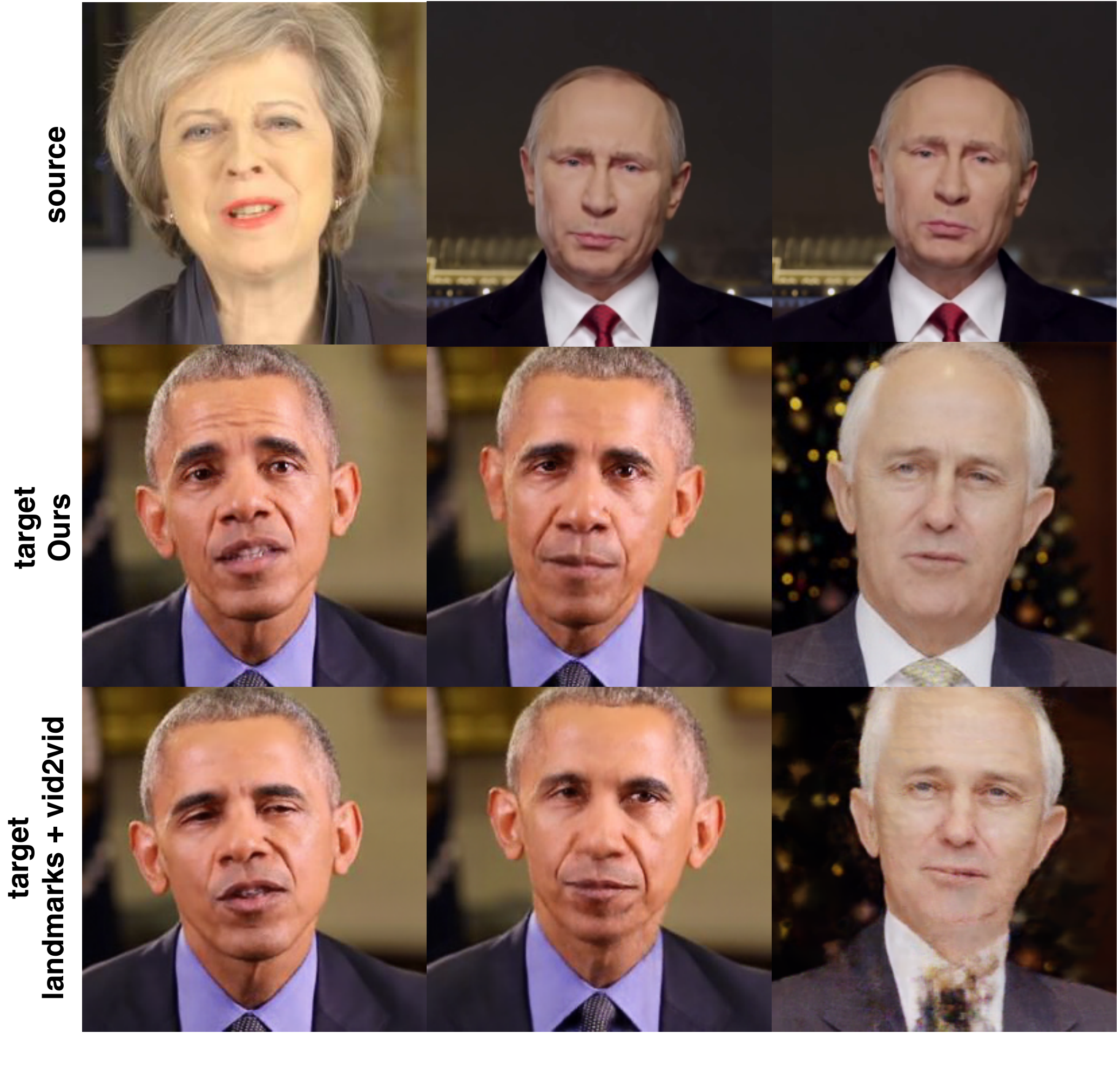}
    \caption{Qualitative comparison with \textit{vid2vid} conditioned on landmarks \cite{vid2vid}. The use of 3DMM and conditioning on $\mathbf{NMFC}$ images enables identity and expression disentanglement. Our neural video renderer preserves the identity of the target, in contrast to \textit{vid2vid}, which distorts it.
    }
    \label{fig:qualitative_3}
\end{figure}

\setcounter{figure}{6}

\subsection{Qualitative Results}

We show that our \textbf{head2head} framework performs a highly accurate transfer of the expression, head pose and eye gaze from the source to the target subject. We further compare our method with the state-of-the-art \textit{Deep Video Portraits} \cite{deepvideoportraits} and two strong baselines: 1) \textit{pix2pix} \cite{pix2pix} conditioned on our novel $\mathbf{NMFC}$ and the eye gaze images and 2) \textit{vid2vid} conditioned on facial landmarks, as in \cite{vid2vid}. 

We trained both baseline methods and our model on two different target sequences (Obama and Turnbull). Then, we performed a self-reenactment experiment for the two target subjects and a full head transfer experiment from May to Obama, Putin to Obama and Putin to Turnbull. We found that the lack of sequential modeling when using an image-to-image transfer method, such as \textit{pix2pix}, results in low quality and temporally incoherent frames. Examples from our self-reenactment experiment on the test set of Obama and Turnbull, are shown in Fig \ref{fig:qualitative_2}. It can be seen that our method produces more convincing results than the pix2pix baseline conditioned on 
$\mathbf{NMFC}$ and eyes images, for the task of video reconstruction. Next, we demonstrate the importance of the 3DMM and our novel $\mathbf{NMFC}$ conditioning input to the neural video renderer. Given a sequence of driving frames, our method successfully disentangles identity and expression and generates a set of target frames with the head movements of the source, while preserving facial attributes of the target identity. On the other hand, \textit{vid2vid} model conditioning on landmarks distorts the appearance of the target, since landmarks contain identity information from the source sequence, which is then passed to the generated target video, as seen in Fig. \ref{fig:qualitative_3}. To make the comparison fairer, we rigidly registered the extracted landmarks of Putin to the target's. This helps to reduce the distortion coming from the difference in scale, otherwise the \textit{vid2vid} network generates totally distorted and nonsensical images.

When comparing with \cite{deepvideoportraits}, our method performs equally well in terms of image quality and photo-realism, while we achieve superior results in the interior of the mouth. As can be seen in Fig. \ref{fig:qualitative_1}, our method outperforms the pose and expression transfer of \cite{deepvideoportraits} in many cases. The head movements and facial expressions of our generated sequence match precisely with the ones in the source video.

\subsection{Ablation Study}

\begin{figure}[h!]
    \centering
    \includegraphics[width=1.0\linewidth]{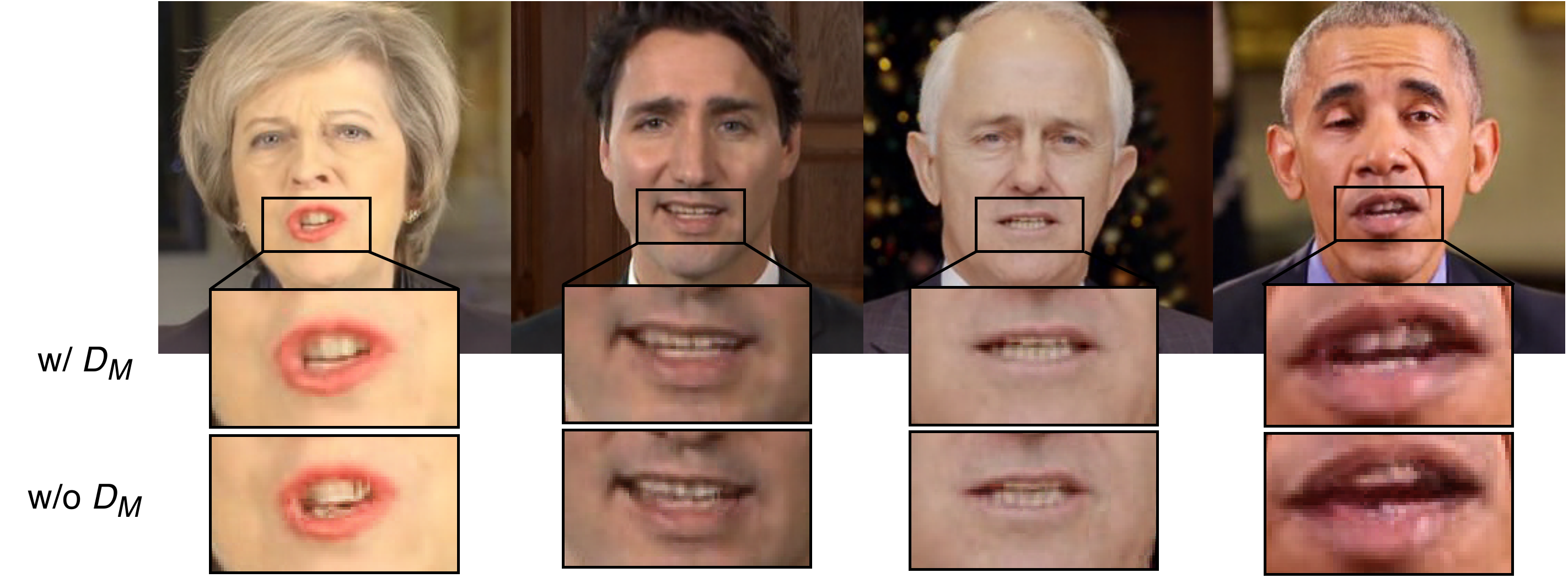}
    \caption{Generated frames by our \textbf{head2head} method, with and without the mouth discriminator $D_M$.}
    \label{fig:abl}
\end{figure}

To evaluate the design choices made to build our \textbf{head2head} framework, we carry out an ablation study demonstrating the effect of each. We first show the effect of combining our framework with FlowNet2 \cite{flownet2} versus our facial flow. Table \ref{tab:ablation_3} (a) reports the average errors achieved when doing the same self-reenactment test in figure \ref{fig:quantitative_1}. We report smaller errors in all cases when using facial flow, which both justifies the significance of utilising our facial flow and makes it more descriptive of the facial dynamics. This is expected, as our flow capitalises on the prior knowledge and was exclusively trained on facial videos. Furthermore, we explore the significance of the mouth discriminator network $D_M$ in the self-reenactment scenario, on four individual target subjects. We measure the average pixel distance in a constrained area around the mouth, between the frames generated with our method and the ground truth. As can be seen in Fig. \ref{fig:quantitative_1} and Fig. \ref{fig:quantitative_2}, mouth cavity is a challenging region, with high errors, mainly due to the absence of conditional input information about the teeth and mouth interiors. The average pixel distances reported in Table \ref{tab:ablation_3} (b) indicate that the use of a mouth discriminator significantly improves our results. The improvements are demonstrated visually in Fig. \ref{fig:abl}.

\begin{table}[h]
  \small
  \centering
  \caption{Ablation study results.}
  \subfloat[Average pixel distance obtained under a self-reenactment setup on the videos in figure \ref{fig:quantitative_1} when combining our method with either FlowNet2 \cite{flownet2} or our facial flow.]{%
    \begin{tabular}{|c|c|c|}
     \hline
      &  \textbf{Ours} ($\downarrow$) & \textbf{Ours} ($\downarrow$) \\
     \textbf{Video} &  \textbf{FlowNet2} &  \textbf{Face-Flow} \\
     \hline
       Justin&  6.9&   6  \\ 
     \hline
       Joe&   9.3&      7.7\\
     \hline
       Merkel&  9 &   7.5 \\
     \hline
       Turnbull&   9.4&     6.5\\
     \hline
    \end{tabular}
  }\hspace{.15cm}%
  \subfloat[Average pixel distance around the mouth area, obtained under a self-reenactment scenario, with and without our dedicated mouth discriminator $D_M$.]{%
    \begin{tabular}{|c|c|c|}
     \hline
     &  \textbf{Ours} ($\downarrow$) &  \textbf{Ours} ($\downarrow$) \\
     \textbf{Video} &  \textbf{ w/o $D_m$} &  \textbf{w/  $D_m$} \\
     \hline
       May & 11.2 & 10.1 \\ 
     \hline
       Justin & 11 & 10.4 \\
     \hline
       Turnbull & 11.3 & 10.7 \\
     \hline
       Obama & 22.3 & 19.4 \\
     \hline
    \end{tabular}
  }
  \label{tab:ablation_3}
\end{table}

%\begin{table}[h!]
%    \centering
%     \caption{Average pixel distance obtained under a self-reenactment setup on the videos in figure \ref{fig:quantitative_1} when combining our method with either FlowNet2 \cite{flownet2} or our facial flow.}
%    \begin{tabular}{|c|c|c|}
%     \hline
%      &  \textbf{Ours} (\downarrow) & \textbf{Ours} (\downarrow) \\
%     \textbf{Video} &  \textbf{$\&$FlowNet2} &  \textbf{$\&$Facial-Flow} \\
%     \hline
%       Justin&  6.9&   6  \\ 
%     \hline
%       Joe&   9.3&      7.7\\
%     \hline
%       Merkel&  9 &   7.5 \\
%     \hline
%       Turnbull&   9.4&     6.5\\
%     \hline
%    \end{tabular}
%    \label{tab:ablation_1}
%\end{table}

%\begin{table}[h!]
%    \centering
%    \caption{Average pixel distance around the mouth area, obtained under a self-reenactment scenario for four different target subjects with and without our dedicated mouth discriminator $D_M$.}
%    \begin{tabular}{|c|c|c|}
%     \hline
%     &  \textbf{Ours} (\downarrow) &  \textbf{Ours} (\downarrow) \\
%     \textbf{Video} &  \textbf{ w/o $D_m$} &  \textbf{w/  $D_m$} \\
%     \hline
%       May & 11.2 & 10.1 \\ 
%     \hline
%       Justin & 11 & 10.4 \\
%     \hline
%       Turnbull & 11.3 & 10.7 \\
%     \hline
%       Obama & 22.3 & 19.4 \\
%     \hline
%    \end{tabular}
%    \label{tab:ablation_2}
%\end{table}

\subsection{Automated Study}
A recent eye-catching attempt in the field to automatically detect fake videos manipulated by state-of-the-art facial manipulation methods was made by Rossler et al. \cite{roessler2019faceforensicspp}. With the help of a well-trained Convolutional Neural Network (CNN), the authors of \cite{roessler2019faceforensicspp} manage to outrank the performance of human observers in detecting manipulated videos. Their dataset (the largest publicly available dataset with facial manipulations) was created by manipulating $1,000$ YouTube videos, depicting real-world scenarios, with graphics-based \cite{face2face, Marek2019} and learning-based \cite{torzdf2019, DeferredNeuralRendering}, facial reenactment methods. In total, their forgery detection network was trained on $1.8$ million facially manipulated frames and reported very high detection accuracy on the test split of their dataset (around $99 \%$).

To assess the realism of our synthesised videos automatically, we utilise the trained forgery-detection network provided by the authors of \cite{roessler2019faceforensicspp}. We randomly choose a subset of 50 videos from their training dataset and manipulate them based on the selected source-target combinations in their work. The accuracy obtained by their network on our generated fake videos is only $1.88 \%$. This demonstrates high photo-realism and consistency of our fake videos, since distinguishing them from real videos is challenging even for such a well-trained system.

\subsection{User Study}

%In addition to the automatic study, we created a small dataset of fake videos and asked the participants to state how realistic each video they just watched looked to them. The set of synthesised videos by our approach embodies different video synthesis scenarios, namely: full-head, self, cycle and facial reenactments, and are of 5-second length each.  This dataset also has videos from 6 different target speakers (May, Merkel, Justin, Turnbull, Putin, a video we received from the authors of \cite{deepvideoportraits}). 

In addition, we performed a user study with two parts. In the first one, we presented to the participants both real and synthesised videos from our method and asked them to assess how realistic they appear. The set of fake videos was generated under three different scenarios, namely: self-reenactment, face-reenactment, full-reenactment. All videos were 5-seconds long. We asked the following question \textit{"On a scale of 1-5, how real does this video look?"}, with 1 meaning "absolutely fake" and 5 "absolutely real". Videos which received a 4 or 5 rating can be considered as "real". The ratings given by 95 anonymous participants are presented in Table \ref{tab:user_study_1} and Table \ref{tab:user_study_2}. The percentage of "real" videos according to the ratings of the participants is comparable for both synthetic and actually real videos. This strongly suggests that our method generates results almost indistinguishable from real videos and indicated that participants were excessively cautious about detecting fake videos, as they had the task explained in advance.

\begin{table}[h!]
\centering
 \caption{Ratings on the self-reenactment task. Columns 1-5 show the number of participants that gave this rating, while column "real" shows the percentage of people that rated the video with a 4 or 5.}
  \begin{tabular}{|c|cccccc|cccccc|}
    \hline
    \multirow{2}{*}{Video} &
    \multicolumn{6}{c|}{\textbf{Our synthesised videos}} &
    \multicolumn{6}{c|}{\textbf{Real videos}} \\
    & 1 & 2 & 3 & 4 & 5 & "real" ($\uparrow$) & 1 & 2 & 3 & 4 & 5 & "real" \\
    \hline
    May & 5 & 8 & 12 & 34 & 36 & \textbf{74\%} & 3 & 6 & 6 & 26 & 54 & 84\%  \\ 
    Turnbull & 11 & 15 & 20 & 29 & 20 & \textbf{52\%} & 2 & 6 & 14 & 40 & 33 & 77\%  \\ 
    Putin* & 4 & 10 & 18 & 36 & 27 & \textbf{66\%} & 2 & 3 & 13 & 37 & 40 & 81\% \\ 
    Justin & 5 & 8 & 18 & 36 & 28 & \textbf{67\%} & 5 & 7 & 11 & 28 & 44 & 76\% \\ 
    dvp* & 3 & 11 & 14 & 26 & 41 & \textbf{71\%} & 9 & 12 & 21 & 29 & 24 & 56\% \\ 
    \hline
    average & 6 & 10 & 16 & 32 & 30 & \textbf{66\% }& 4 & 7 & 13 & 32 & 39 & \textbf{75\%} \\ 
    \hline
  \end{tabular}
  \label{tab:user_study_1}
\end{table}

\begin{table}[h!]
\centering
 \caption{Ratings on the face-reenactment and full-reenactment tasks.}
  \begin{tabular}{|c|cccccc|cccccc|}
    \hline
    \multirow{2}{*}{Video} &
    \multicolumn{6}{c|}{\textbf{Our synthesised videos}} &
    \multicolumn{6}{c|}{\textbf{Real videos}} \\
    & 1 & 2 & 3 & 4 & 5 & "real" ($\uparrow$) & 1 & 2 & 3 & 4 & 5 & "real" \\
    \hline
    Merkel (face-reenact.) & 16 & 21 & 20 & 22 & 16 & \textbf{40\%} & 1 & 6 & 17 & 31 & 40 & \textbf{75\%} \\
    \hline
    Justin (full-reenact.)& 8 & 12 & 25 & 39 & 11 & \textbf{53\%} & 5 & 7 & 11 & 28 & 44 & 76\% \\ 
    dvp* (full-reenact.)& 37 & 19 & 10 & 22 & 7 & \textbf{31\%} & 9 & 12 & 21 & 29 & 24 & 56\% \\ 
    \hline
    average (full-reenact.)& 22 & 16 & 18 & 30 & 9 & \textbf{42\%} & 7 & 10 & 16 & 28 & 34 & \textbf{66\%} \\
    \hline
  \end{tabular}
  \label{tab:user_study_2}
\end{table}

As a second experiment, we presented a source (driving) video along with two target videos, one generated by our \textbf{head2head} method and the other one synthesised by \cite{deepvideoportraits}. Both fake videos were created under a full-reenactment setting, with the source and target subjects being those shown in Fig. \ref{fig:qualitative_1}. Then, we asked the question: \textit{"Which video follows the movements of the person in the source better?"}. Out of the 70 anonymous participants that answered this question, \textbf{60\% have chosen the video produced by our method}, showing an indisputable preference towards our result.

\section{Conclusion}

We presented \textbf{head2head}, a full-reenactment approach for transferring the expression, head pose and the eye gaze from a source to a target actor. Different to the state-of-the-art, the proposed approach produces results that are temporally consistent and capitalises on priors on dense face flow. Furthermore, we pay particular attention to the mouth region in order to further improve quality and realism of the result. We demonstrate that our method can transfer facial motion and head movement more accurately than the state-of-the-art. 

\section{Acknowledgements}
S. Zafeiriou acknowledges funding from a Google Faculty Award, as well as from the EPSRC Fellowship DEFORM: Large Scale Shape Analysis of
Deformable Models of Humans (EP/S010203/1).

%\cite{deepvideoportraits,vid2vid,face2face,DeferredNeuralRendering,fewshot, X2Face, pix2pix}

%\cite{thies2015realtime, Liu2001, Suwajanakorn2017, Olszewski2017, Vlasic2005, elor2017bringingPortraits,Garrido2014,Garrido2015}

%\cite{ganGoodfellow, cgan, PGAN, Wang2018, CycleGAN2017}

%\cite{Chen_2017_ICCV, Denton2017, Finn2016, Saito2017TemporalGA, VondrickNIPS2016, tulyakov2017mocogan, UNIT, TaigmanPW16, lsgan}
\vspace{-0.3cm}
\bibliographystyle{ieeetr}
\bibliography{bibliography}

\end{document}